\documentclass{article}
\usepackage{icassp/spconf}

\usepackage{booktabs}                  
\usepackage{multirow}
\usepackage{spconf,amsmath,graphicx,hyperref}
\usepackage{subcaption}

\title{A State-Update Prompting Strategy for Efficient and Robust Multi-turn Dialogue}

\name{Ziyi Liu\thanks{The author acknowledges computational support from The Hong Kong University of Science and Technology and The Hong Kong University of Science and Technology (Guangzhou); and is grateful to a professor at City University of Hong Kong for their insightful guidance.}}
\address{
  School of Artificial Intelligence \\ 
  Beijing University of Posts and Telecommunications \\ 
  \texttt{ZiyiLiu0811@outlook.com} 
}

\begin{document}

\maketitle
\ninept 
\begin{abstract}
Large Language Models (LLMs) struggle with information forgetting and inefficiency in long-horizon, multi-turn dialogues. To address this, we propose a training-free prompt engineering method, the State-Update Multi-turn Dialogue Strategy. It utilizes "State Reconstruction" and "History Remind" mechanisms to effectively manage dialogue history. Our strategy shows strong performance across multiple multi-hop QA datasets. For instance, on the HotpotQA dataset, it improves the core information filtering score by 32.6\%, leading to a 14.1\% increase in the downstream QA score, while also reducing inference time by 73.1\% and token consumption by 59.4\%. Ablation studies confirm the pivotal roles of both components. Our work offers an effective solution for optimizing LLMs in long-range interactions, providing new insights for developing more robust Agents.
\end{abstract}

\section{Introduction}
\label{sec:introduction}
Large Language Models (LLMs) have demonstrated remarkable, human-like capabilities across a vast spectrum of tasks\cite{radford2019language-models}\cite{brown_language_2020}\cite{ouyang_training_2022}, from complex reasoning to fluent text generation.Efforts to advance LLMs and address their inherent flaws, such as hallucination, have pursued several key strategies. Prompt engineering techniques, exemplified by Chain-of-Thought (CoT)\cite{wei_chain--thought_2022}, aim to refine a model's internal reasoning by promoting step-by-step deliberation. Concurrently, methods like Retrieval-Augmented Generation (RAG)\cite{ragforknoledge-intrnsivetasks}\cite{gao2024retrievalaugmentedgenerationlargelanguage} and tool-using\cite{toolformer}\cite{yao2023react}\cite{liu2025longvideoagentmultiagentreasoninglong} enhance factual accuracy by integrating external knowledge from databases and search engines\cite{jin2025searchr1}. Although these approaches differ, they share a critical prerequisite: efficient information filtering. Whether processing complex internal thought processes or large volumes of external documents, an LLM's ability to distill relevant information is paramount for effective downstream reasoning. This dependency has established information filtering as a central bottleneck for developing more capable and reliable LLM systems.

To tackle this filtering challenge, information can be presented to an LLM through two primary interaction paradigms: \textit{single-turn input}, where all information is contained within a single prompt, preserving complete information presentation and internal relationships; and \textit{multi-turn input}, where information is segmented into multiple subsets and progressively input through multiple dialogue turns, allowing the model to focus on shorter contexts with potentially greater reasoning space. However, existing multi-turn dialogue mechanisms exhibit notable limitations, a concern that has been echoed in recent research. demonstrated that LLMs suffer significant performance degradation in multi-turn conversations, with models literally "getting lost" as dialogue progresses\cite{laban2025llmslostmultiturnconversation}. This aligns with our observations that multi-turn interactions, while theoretically offering advantages in context management, face fundamental challenges in practice. First, models demonstrate significantly better comprehension of recently input information compared to earlier inputs, exhibiting a "forgetting phenomenon"\cite{liu-etal-2024-lost}.Second, LLMs struggle to effectively integrate information across different turns for holistic reasoning.

\begin{figure*}[t!]
    \centering
    \begin{subfigure}[b]{0.48\linewidth}
        \centering
        \includegraphics[width=0.6\linewidth]{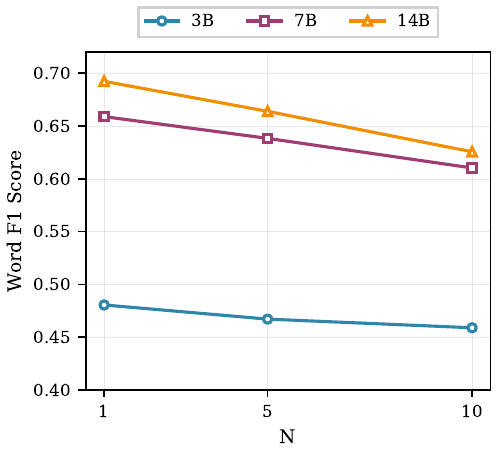}
        \caption{Performance degradation with increasing dialogue turns}
        \label{fig:experiments_a}
    \end{subfigure}
    \hfill
    \begin{subfigure}[b]{0.48\linewidth}
        \centering
        \includegraphics[width=0.6\linewidth]{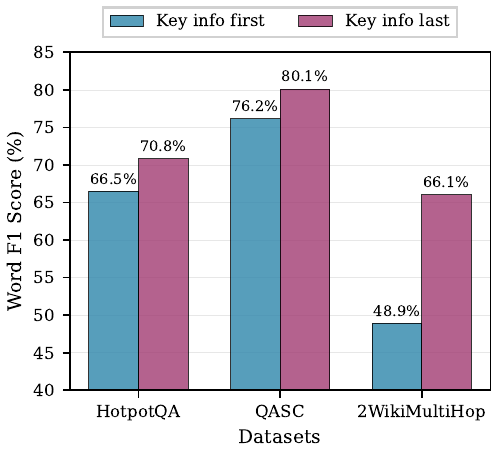}
        \caption{Impact of information position on comprehension}
        \label{fig:experiments_b}
    \end{subfigure}
    
    \caption{Experimental validation results. In (a), F1 scores decrease as the number of turns increases, with larger models showing more pronounced degradation on HotpotQA dataset. In (b), models demonstrate better performance when key information is placed in later turns.}
    \label{fig:experiments}
\end{figure*}

To empirically validate these hypotheses, we designed two targeted experiments using the HotpotQA benchmark\cite{yang-etal-2018-hotpotqa}. 
First, we investigated the impact of dialogue length on performance. 
We segmented the 10 provided paragraphs into varying numbers of conversational turns ($N=1, 5, 10$) and tasked the model with filtering key information at each step. 
As shown in Figure~\ref{fig:experiments_a}, F1 scores consistently decrease as the number of turns increases. 
Notably, this performance degradation is more pronounced in larger models, suggesting that while scaling enhances single-turn comprehension, it does not resolve the challenge of retaining historical context in multi-turn dialogues. 
Furthermore, to diagnose the cause of this degradation, we examined the impact of information position. 
We systematically varied the placement of crucial paragraphs, comparing scenarios where they appeared in the first versus the last turn. 
The results, presented in Figure~\ref{fig:experiments_b}, show that placing key information in the final turn yields significantly higher performance. 
This finding provides strong evidence for a pronounced recency bias, or ``forgetting phenomenon,'' where the model fails to effectively utilize information from earlier turns.

To address these limitations, we propose a training-free prompt engineering approach called the \textbf{State-Update Multi-turn Dialogue Strategy}. This strategy comprises three core technical components: (1) \textit{State Reconstruction}, where each dialogue turn does not retain complete dialogue history but reconstructs the dialogue state to reduce token consumption; (2) \textit{History Reminder}, which explicitly reminds the model of previously identified key information through a ``Previously selected'' mechanism; and (3) \textit{XML Structured Output}, using \texttt{<info>} tags to ensure result parsability and consistency. Through this design, the strategy improves information filtering performance while significantly reducing token consumption, effectively addressing the problems of insufficient information utilization and forgetting phenomena in traditional multi-turn dialogues.

The main contributions of this work can be summarized as follows:
\begin{itemize}
    \item We systematically identify and quantitatively analyze core challenges in LLM multi-turn dialogues, including insufficient cross-turn information utilization, inadequate reasoning coherence, and attention bias toward recent inputs.
    
    \item We propose a simple yet effective state-update multi-turn dialogue strategy that improves model performance while significanvtly reducing token consumption, offering dual advantages in both time and space efficiency.
    
    \item Through comprehensive experimental validation, we demonstrate the effectiveness of our method and elaborate on its potential applications in agent systems and RAG technologies, providing new research directions for related fields.
\end{itemize}

\section{Related Work}
\label{sec:related}
A central strategy for improving the performance of  LLMs is to equip them with high-quality context. Current research has largely proceeded along two lines of inquiry. The first centers on incorporating external knowledge, epitomized by frameworks such as Retrieval-Augmented Generation (RAG)\cite{ragforknoledge-intrnsivetasks}\cite{gao2024retrievalaugmentedgenerationlargelanguage} and Tool-using\cite{toolformer}\cite{yao2023react}\cite{jin2025searchr1}. These methods provide models with the factual grounding necessary for generating responses by interfacing with external knowledge bases or search engines, primarily tackling the challenge of knowledge acquisition and injection. However, such methods face challenges, including the retrieval of irrelevant information\cite{yue2025inference-rag}\cite{zhao2022densetextretrievalbased} and the failure to supply sufficiently useful context\cite{jiang-etal-2023-active-rag}\cite{jin2025longcontext}. The efficacy of tool-use, in particular, is highly dependent on the quality of the retrieved information.

\begin{figure*}[t!]
    \centering
    \includegraphics[width=0.9\linewidth]{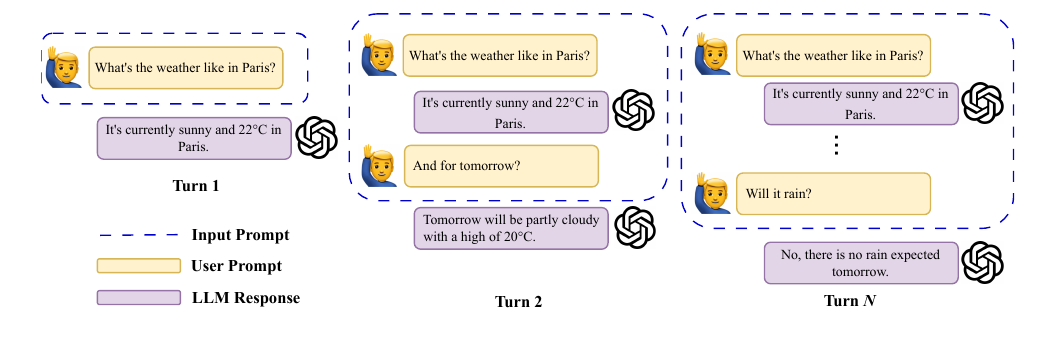}
    \caption{Traditional multi-turn dialogue strategy baseline approach.}
    \label{fig:baseline}
\end{figure*}

In this work, we explore a complementary avenue: the dynamic management and efficient utilization of the model's internal information flow. When multi-turn dialogue is required, the most common strategy is linear history concatenation\cite{sordoni_neural_2015-conversation}\cite{brown_language_2020}\cite{ouyang_training_2022}\cite{touvron2023llama2openfoundation}, a process illustrated in Figure~\ref{fig:baseline}. However, this approach has been demonstrated to lead to the "Lost in the Middle" phenomenon—a tendency for models to over-rely on information at the extremities of the context while neglecting the intermediate parts, which compromises conversational coherence. Consequently, our work is focused on optimizing LLM performance in multi-turn dialogue settings.

\section{Method}
\label{sec:method}

\begin{figure*}[t!]
    \centering
    \includegraphics[width=0.9\linewidth]{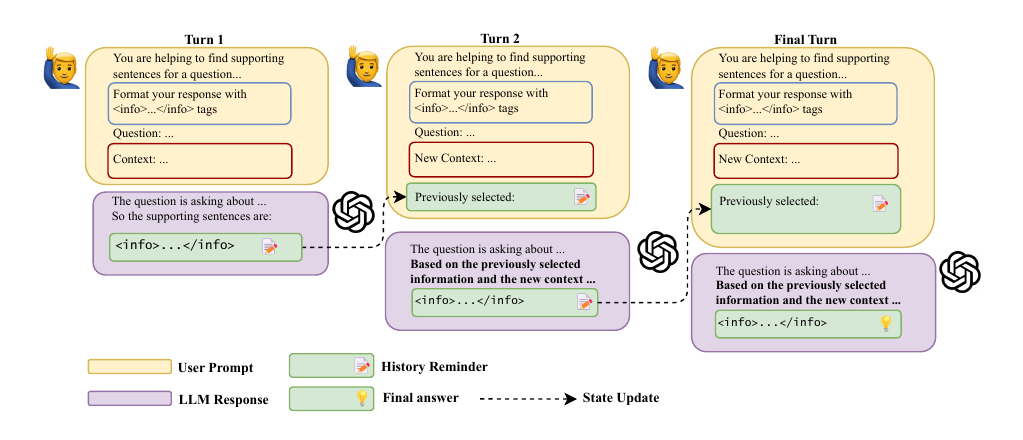}
    \caption{Overview of our proposed State-Update Multi-turn Dialogue Strategy framework with three core components: State Reconstruction, History Reminder, and XML Structured Output.}
    \label{fig:overview}
\end{figure*}
\label{sec:method}

\begin{figure*}[t!]
    \centering 

    \begin{subfigure}[b]{0.39\textwidth}
        \centering
        \includegraphics[width=\linewidth]{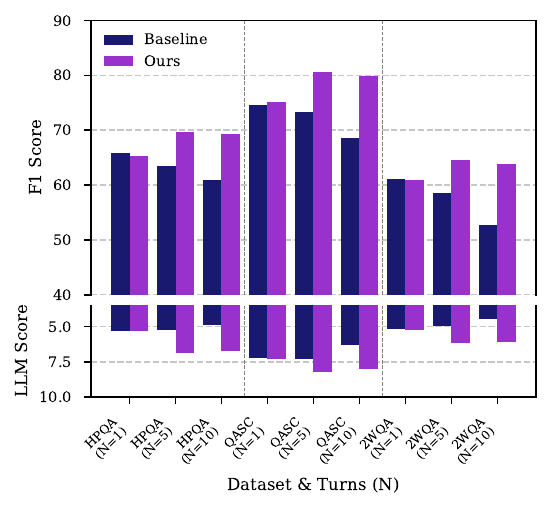}
        \caption{Performance Analysis} 
        \label{fig:perf_analysis}
    \end{subfigure}
    \hfill 
    \begin{subfigure}[b]{0.6\textwidth}
        \centering
        \includegraphics[width=\linewidth]{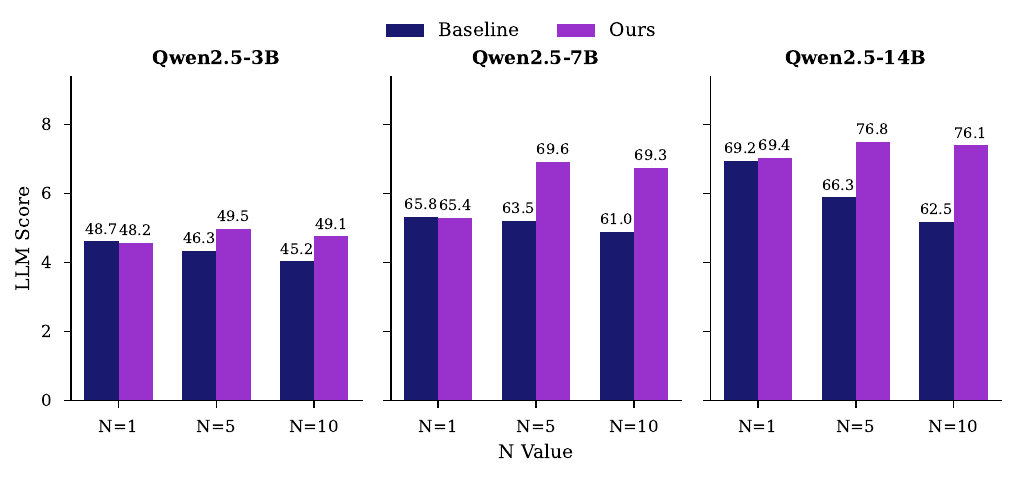}
        \caption{Impact of Model Scaling}
        \label{fig:model_scaling}
    \end{subfigure}

    \caption{Comprehensive performance evaluation. 
    (a) F1 Score comparison of our method against the baseline on three multi-turn QA datasets (HotpotQA, QASC, and 2WikiMultiHopQA) with varying conversation lengths ($N=1, 5, 10$). Our state-update strategy consistently and significantly outperforms the baseline, especially in longer dialogues ($N>1$), demonstrating its robustness against the degradation of long-context information.
    (b) An analysis of performance as a function of model scale, comparing our method with the baseline across different model sizes and $N$ values.
    }

    \label{fig:combined_results}
\end{figure*}
To address the challenges of positional bias and information degradation in traditional multi-turn dialogues, we propose a training-free prompt engineering approach named the \textbf{State-Update Multi-turn Dialogue Strategy}. This strategy replaces the conventional method of linearly appending conversation history by reconstructing the dialogue state at each turn. This allows for the reliable accumulation of information while maintaining an efficient, fixed-size context window.
The implementation of this strategy relies on a meticulously designed prompt architecture:

\begin{center}
\noindent 
\fbox{
    \begin{minipage}{0.9\linewidth} 
    \small
    \textbf{SYSTEM:} You are helping to find supporting sentences... Format your response with \textless info\textgreater...\textless/info\textgreater tags. Each response should contain ALL supporting sentences (previous + new ones).
    \end{minipage}
}
\end{center}
A unified system prompt is employed across all turns to establish two core tasks for the model: (1) \textbf{Structured Output}: The model is required to encapsulate all supporting sentences within XML tags (\textless info\textgreater) to ensure reliable and parsable outputs. (2) \textbf{Cumulative Principle}: The model is instructed that each response must contain \textbf{all} supporting sentences found in previous turns as well as the current one.

The conversation proceeds as follows:
\begin{itemize}
    \item \textbf{First Turn (Initialization):} The user provides the question and the initial text passage to start the identification process.
    
    \begin{center}
    \noindent
    \fbox{
        \begin{minipage}{0.9\linewidth}
        \small
        \textbf{USER:} Question: ...
        
        Context: \{passage\_1\}
        
        \end{minipage}
    }
    \end{center}

    \item \textbf{Subsequent Turns ($k \geq 2$, State Update):} The dialogue state is reconstructed. The user input includes the new text passage, supplemented by the supporting sentences extracted from the previous turn, which serve as an "explicit history reminder."

    \begin{center}
    \noindent
    \fbox{
        \begin{minipage}{0.9\linewidth}
        \small
        \textbf{USER:} 
        
        Question: ...
        
        New Context : \{passage\_k\}

        \vspace{0.5mm} 
        
        \textbf{USER:} Previously selected: \{supporting sentences from turn k-1\}
        \end{minipage}
    }
    \end{center}
\end{itemize}

The core mechanism of this flow is that after each turn, we parse the model's output within the \textless info\textgreater tags and use it as the content for the "Previously selected" field in the subsequent turn. This design offers two significant advantages:
\begin{itemize}
    \item \textbf{Efficiency Enhancement:} Compared to linear history concatenation, the state reconstruction approach significantly reduces the number of input tokens per turn, thereby lowering computational costs and inference latency.
    \item \textbf{Mitigation of Forgetting:} By re-injecting key historical information as an explicit reminder, the strategy compels the model to attend to and integrate the entire context. 
\end{itemize}

\section{Experiments}
\label{sec:experiments}


\label{sec:experiments}

\subsection{Experimental Setup}

\noindent\textbf{Datasets.} We evaluate our method on three public benchmarks: \textit{HotpotQA}, \textit{2WikiMultiHopQA}, and \textit{QASC}. \textit{HotpotQA} and \textit{2WikiMultiHopQA}\cite{xanh2020_2wikimultihop} are multi-hop QA datasets that require reasoning over contexts containing distractor information. For \textit{QASC}\cite{Khot2019QASC}, a non-reasoning QA dataset, we construct a similar multi-turn format by randomly sampling sentences to form the context, mirroring the setup of the other datasets.

\noindent\textbf{Models.} All experiments are conducted on the Qwen2.5-Instruct series of models\cite{qwen2025qwen25technicalreport}, using a consistent decoding temperature of 0.8 and default values for other key hyperparameters.

\noindent\textbf{Baseline.} We compare against the standard multi-turn conversation strategy as our baseline. In this approach, the model context for each turn is formed by concatenating the current input with the entire preceding conversation history.

\noindent\textbf{Metrics.} We employ two metrics to assess the quality of the extracted supporting sentences: \textbf{Word F1 Score}, a token-overlap-based metric, and an \textbf{LLM-based Score} (Info Score and QA Score) evaluated by Gemini 2.5 pro\cite{comanici2025gemini25pushingfrontier} to capture semantic quality.

\begin{table}[t] 
\centering
\caption{Main results and ablation study. Our method outperforms the baseline in both effectiveness (Info/QA Score) and efficiency (Time/Tokens). Ablation studies demonstrate the necessity of both the HR and SR components. (↑: Higher is better, ↓: Lower is better)}
\label{tab:ablation}
\resizebox{\columnwidth}{!}{%
\begin{tabular}{@{}lccccc@{}}
\toprule
\textbf{Method} & \textbf{Info Score}↑ & \textbf{Time (s)}↓ & \textbf{Tokens}↓ & \textbf{QA Score}↑ \\
\midrule
Baseline        & 5.21                   & 47.9               & 6707             & 6.33 \\
Ours            & \textbf{6.91}          & \textbf{12.9}      & \textbf{2721}    & \textbf{7.22} \\
\midrule
\multicolumn{5}{@{}l}{\textit{Ablation Study}} \\
\quad w/o HR    & 3.74                   & 10.8               & 2514             & 4.68    \\
\quad w/o SR    & 6.71                   & 45.2               & 6526             & 6.41    \\
\bottomrule
\end{tabular}%
}
\end{table}
\subsection{Results and Analysis}

Our State-Update strategy significantly outperforms the baseline in mitigating catastrophic forgetting during long-turn dialogues. As illustrated in Figure~\ref{fig:perf_analysis}, the baseline's performance degrades sharply as the conversation progresses, whereas our method maintains a robust performance trend. At $N=10$ turns, our approach achieves an average Word F1 improvement of approximately 10\% and an Info Score increase of over 1.5 points, demonstrating its ability to effectively integrate historical information and alleviate the model's forgetting problem.

Furthermore, our strategy offers substantial computational efficiency. As detailed in Table~\ref{tab:ablation}, in conversations of $N=5$ turns, our method reduces total token consumption by \textbf{59.4\%} and inference time by \textbf{73.1\%}. This gain stems from its ability to avoid re-processing the entire conversation history at each turn, establishing it as a highly efficient and practical solution.

To validate the robustness and generalizability of our method, we evaluated its performance across diverse datasets and model scales. Our approach consistently outperforms the baseline on all three datasets, including \textit{HotpotQA}, \textit{2WikiMultopQA}, and \textit{QASC} (Figure~\ref{fig:perf_analysis}). Moreover, this advantage holds across model sizes ranging from 3B to 14B parameters (Figure~\ref{fig:model_scaling}), underscoring that our strategy is a \textbf{fundamental improvement} over simple history concatenation, rather than an artifact of a specific experimental setup.

Crucially, the enhanced information filtering directly translates to superior performance on downstream tasks. As shown in Table~\ref{tab:ablation}, the context distilled by our method enables the model to achieve a \textbf{14.1\%} higher score on the final question-answering task compared to the baseline. This confirms that the preserved dialogue state is not only compact and efficient but also highly valuable for complex reasoning, firmly establishing its \textbf{significant practical impact}.

\subsection{Ablation Study}
To validate the architectural design of our method, we conducted an ablation study (Table~\ref{tab:ablation}) on its two core components: State Reconstruction (SR) and History Reminder (HR). Removing the SR module (\texttt{w/o SR}) caused efficiency to plummet to near-baseline levels, confirming its primary role in context compression. Conversely, removing the HR module (\texttt{w/o HR}) led to a catastrophic performance drop, with the Info Score falling from 6.91 to 3.74 and the QA Score from 7.22 to 4.68. This result strongly validates that HR is the cornerstone for cross-turn reasoning and contextual understanding. The study confirms that both components are indispensable, with SR ensuring efficiency and HR driving performance.

\section{Conclusion}
\label{sec:conclusion}
We introduce a novel, training-free State-Updating Multi-turn Dialogue Strategy that leverages a state reconstruction and a history-aware reminding mechanism. Our approach not only significantly enhances performance on information filtering and downstream question-answering tasks by 14.1\% but also drastically reduces computational overhead, achieving a 73.1\% reduction in inference time and a 59.4\% decrease in token consumption. This research confirms that actively and explicitly managing dialogue state is a more efficient interaction paradigm compared to conventional history concatenation methods, offering crucial insights for building more robust memory modules for intelligent agents. As a future direction, we propose exploring how to better manage context input through prompt engineering to make better use of LLMs.

\vfill\pagebreak
\bibliographystyle{icassp/IEEEbib}
\bibliography{ref}

\end{document}